# Deep learning, machine vision in agriculture in 2021

Rakhmatulin Ildar, PhD

*South Ural State University, Department of Power Plants Networks and Systems, Chelyabinsk city, Russia, 454080*

ildar.o2010@yandex.ru

## Abstract

Unprecedented progress in the deep learning field influenced many of different industries, including agriculture sector. The use of neural networks in agro-industrial activity in the task of recognizing cultivated crops and weed is a new direction with a history lower than 10 years. Dozens of new neural networks appear every year, but absence of any standards significantly complicates the understanding of the real situation of the use of the neural network in the agricultural sector. In the manuscript, we analyzed research over the past 10 years on the use of neural networks for the classification and tracking of crops and weeds in agriculture. We presented the analysis of the results of using various neural network algorithms for the task of classification and tracking. Finally, we made recommendations for the use of neural networks in the tasks of recognizing a cultivated object and weeds.

## 1. Introduction

The use of artificial intelligence has expanded rapidly in recent years. Researchers from various fields of science use in practice the functionality of neural networks of machine learning and machine vision. But neural networks have very advanced functionality and variety, and if a researcher has not been directly involved in working with neural networks over the past years, then it is very difficult for him at the initial stage to efficiently select the correct model of a neural network. Therefore, the task of this manuscript provides information on current trends in the field of neural networks and weed detection field.

We considered only completed studies in which presented the test results in neural networks and weed controls system According to their characteristics, these devices should have been suitable for the field application. Search for articles by keywords was carried out in the following publishers: Elsiver, Taylor & Francis, Springer, Wiley, Informa. Keyword searches were also done on the Google search engine and scholar.google.com for the last 10 years.

The following papers are similar in content to this manuscript. Ip et al. [1] provided an overview of significant research in plant protection using big data with an emphasis on weed control and management. Authors considered machine learning methods for big data analysis, which is not suitable for every type of research. Wang et al [2] presented a similar paper, where summarized the achievements in detecting weeds using ground-based methods of machine vision and image processing. These manuscripts describe studies without any systematic approach, according to various criteria, because of which it is quite difficult to imagine a complete picture of the use of neural networks in the agricultural sector.

Depending on weed control methods for a simple understanding, we divided considered the manuscript into several subgroups, figure 1.

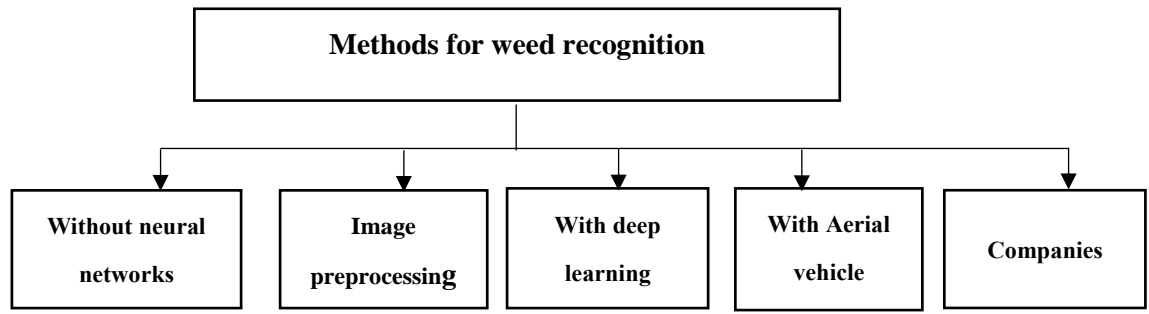

Fig.1. Classification of the manuscript on the use of neural networks in the tasks of detecting weeds

## 2.1. Weed recognition without neural networks

The task of recognizing crops and weeds already faced farmers decades earlier. As a result of which there are many types of research that, in the context of neural networks, have now lost their former relevance, but at the same time, due to innovative technical ideas and non-standard mathematical solutions, they carry a certain scientific value. Nejati et al. [3], to determine weed in carrots used the active stereoscopy method based on coded structured light with time multiplexing, designed to consider the features of a small-scale scene. This is a rather expensive method and can only be used to compare the results with the results obtained from neural networks. Nejati et al. [4] used the process of identifying a weed for a fast Fourier transform to convert the density of the leaf edge. This method can be implemented as an auxiliary function in the process of preparing images for neural networks.

To some extent, works may be useful that is directly not related to the subject of machine vision, but of interest, due to the presence of physical actuators that can be used as part of devices with machine vision. Rueda-Ayala et. al [5], Utstumo et al. [6], Alba et al [7], Melander et al. [8] and Syrov [9] Representative devices for mechanical weed control, which can be easily connected with various systems based on neural networks.

## 2.2. Image preprocessing in weed recognition tasks

The quality of neural network operation directly depends on image pre-processing. Therefore, this aspect is one of the most important in the field of machine vision. There are no direct criteria indicating how to process the image processing since each case is unique and requires its own special method.

Panqueba et al. [10], used image segmentation function and the use of a neural network trained with sigmoid activation function type. They implemented script that automatically processes images before they are sent to the input of a neural network. As a result, a rather complicated computational process was used to obtain a positive result which can significantly increase network calculation time. But Reddy et al. [11] for similar tasks managed to achieve a positive result only with filters such as Gaussian and Laplacian. It is not possible to compare these methods on the author's datasets since they are not presented in the public domain.

Vikhram et al. [12] and Pasha et. al [13] used cv2.inRange() function by OpenCV library in the weed tracking tasks. This is a very simple and popular method that works with high precision in laboratory conditions. That is, in conditions where can specifically create the necessary color contrast.

Many works in which the authors do not use popular computer vision libraries to implement the image processing process. These works involve the implementation of image processing due to mathematical formulas based on Matlab or other software. For example, García-Santillán et al. [14], implemented image segmentation process based on the use of the vegetation index and linear least squares regression to correct the line. Murawwat [16] used binary classification to prepare images before start uses neural networks. Raja et al. [18] developed software for image processing. Software realized by the algorithm is based on erosion followed by a dilatation segmentation algorithm. But in these research, verification with other methods was not carried out. To repeat the experiment, the installation of the software is required. Moreover, these methods are extremely difficult to implement in conjunction with neural networks. It is necessary to use a stationary computer with sufficient power of GPU for the implementation of calculations. Unlike these works, Murawwat et al. [15] used of machine learning procedures for the implementation of reference vector machines (SVMs) and BLOB objects for crop detection task with low power CPU. Bhongal et al. [17] used similar standard functions and used segmentation and subsequent use of a neural network for image processing. Ambika et al [19] and Ferreira et al. [20] focused their research on the analysis of modern image processing methods.

A common drawback of the works described above is the lack of a complete literature review, which is why there are many duplicating works. In such works, the process of image preprocessing is similarly described, or a similar model of a neural network with the same neuron activation function is used. It is worth noting that the geography of research is extensive, both in the countries of researchers and in crops and weeds. To verify recognition, the works cite artificially designed mock-ups with a pronounced color contrast, which makes it impossible to evaluate how these models work in practice. These works can be considered as the implementation of various methods of machine learning and image preprocessing in practice, using concrete crops and weeds as an example.

A common feature of these works, and to a lesser degree subsequent one, is an overly detailed description of image preprocessing, which, although it is an important component of research related to machine vision, it still does not carry scientific novelty. At the same time, assessing the correctness of preprocessing in the described works is a difficult task, since each case is unique and depends on the tasks and the number of photos submitted for processing.

According to the analysis, we can conclude that for the most part, not all the possible functionality for preprocessing images used, for example, that can be used in the OpenCV library. The following is a list that is useful for preprocessing images. In parentheses are the operators used to call an object in the python programming language in the OpenCV library.

Image predprocessing:

**- Color;**

    - color filters (cv2.createTrackbar, cv2.cghhnbnmlootroiesxxxvtColor, color_image[:,:,n])

    - color detection (cv2.inRange, cv2.findContours);

**- Smoothing Images**:

        - Convolution (cv2.filter2D);

        - Image Blurring (cv2.blur);

        - Gaussian Filtering (cv2.GaussianBlur);

        - Median Filtering (cv2.medianBlur);

        - Bilateral Filtering (cv2.bilateralFilter);

        - Laplacian (cv2.Laplacian).

**- Morphological Transformations:**

        - Erosion (cv2.erode);

        - Dilation (cv2.dilate);

        - Opening (cv2.MORPH_OPEN);

        - Closing (cv2.MORPH_CLOSE);

        - Morphological Gradient (cv2.MORPH_GRADIENT);

        - Top Hat (cv2.MORPH_OPEN);

        - Black Hat (cv2.MORPH_BLACKHAT).

**Edge Detection (**cv2.Canny**).**

**Contours:**

    - Contours **(**cv.drawContours**);**

    - Shapes (cv2.rectangle, cv2.minEnclosingCircle, cv2.circle, cv2.line);

    - Fill (cv2.floodFill).

**Thresholding:**

        - Binary (cv2.THRESH_BINARY);

        - Binary_inv (cv2.THRESH_BINARY_INV);

        - Trunc (cv2.THRESH_TRUNC);

        - Tozero (cv2.THRESH_TOZERO);

        - Tozero_inv (cv2.THRESH_TOZERO_INV).

**Geometrical image transformation:**

- Resize (cv2.resize);

- Translation cv2.warpAffine();

- Rotation (cv2.getRotationMatrix2D);

- Affine Transformation (cv2.getAffineTransform);

- Perspective Transformation (cv2.getPerspectiveTransform, cv2.warpPerspectiv).

**Image Derivatives:**

- Sobel derivatives (cv2.Sobel);

- Scharr derivatives (cv2.Scharr).

Using these functions will allow you to evaluate and understand what elements in the image may be helpful for weed recognition. For example, in figure 2, we realized the most popular functions for the image recognition process (https://github.com/Ildaron/OpenCV-image-preprocessing-python). The separation of the plant on the leading edge from the background presents certain difficulties. The solution is to select the correct method and select the correct coefficients of threshold function.

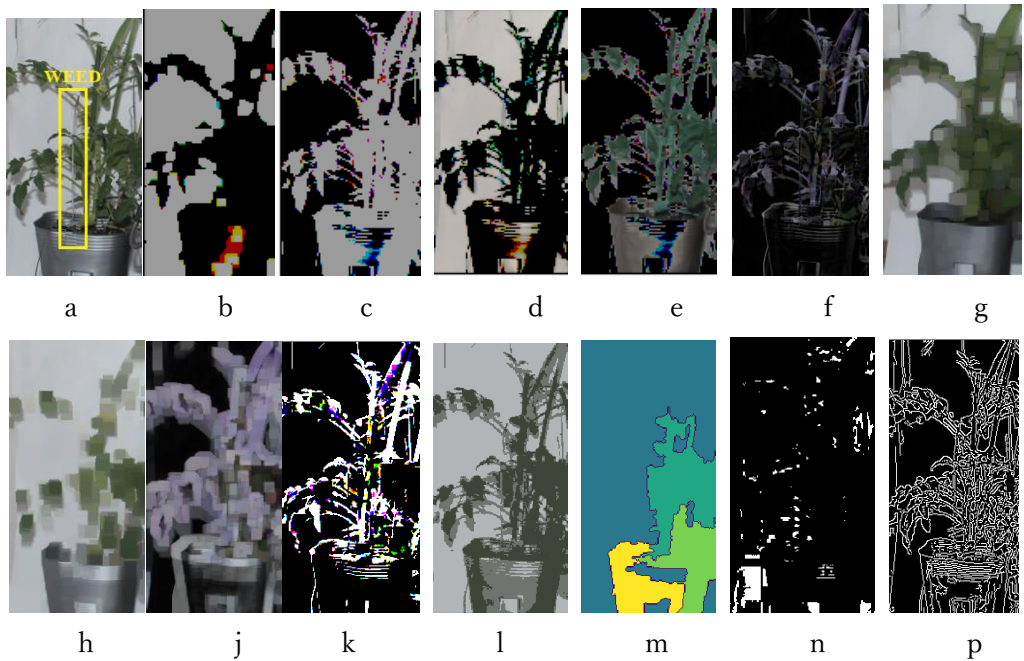

Fig.2. Feature extraction: a - received image with weed and tomato, b - image thresholding by cv2.threshold(img,127,155,cv2.THRESH_BINARY), c - image thresholding by cv2.threshold(img,127,155,cv2.THRESH_BINARY_INV), d - image thresholding by cv2.threshold(img,127,155,cv2.THRESH_TOZERO), e - image thresholding by cv2.threshold(img,127,155,cv2.THRESH_TOZERO_INV), f - morphological transformations by cv2.morphologyEx(img, cv2.MORPH_BLACKHAT, kernel1), g - morphological transformations by cv2.morphologyEx(img, cv2.MORPH_OPEN, kernel2), h - morphological transformations by cv2.morphologyEx(img, cv2.MORPH_CLOSE, kernel2), j - morphological transformations by cv2.morphologyEx(img, cv2.MORPH_GRADIENT, kernel2), k -flood fill by cv2.floodFill(im_floodfill, mask, (0, 0), 255), l - segmentation by cv2.kmeans(pixel_values, k, None,

criteria, 10, cv2.KMEANS_RANDOM_CENTERS), m - distance transform by cv2.distanceTransform(opening,cv2.DIST_L2,5), n - color detection by cv2.inRange(hsv, h_min, h_max) – from

## 2.3. Deep learning in weed recognition tasks

In recent years, deep learning in connection with the availability of computing power began to be widely used in machine vision problems. Today, the use of deep machine learning is a priority in the problems of classification and tracking, which is confirmed by the results of competitions at Kaggle (www.kaggle.com) and Image.net. The most popular neural network used in classification tasks is the convolutional neural network (CNN). In the manuscripts, the authors use various models of neural networks for the classification of weeds and crop tasks. As is correct, to confirm the effectiveness of the proposed methods, the authors give comparisons of the proposed model with other models of neural networks. But each author decides on the success of the proposed method in the manuscript himself without any standards. For example the next papers used different models of neural networks for weed recognition tasks. Bah et al. [32] considered both the CNN network and the residual CNN for image classification. Priya et al. [33] created the deep neural network using the Keras library. Potena et al. [34] used the CNN neural network for pixel segmentation of a binary image, for extracting pixels and constructing a projection of three-dimensional points. Ahmed et al [35], created the network is based on VGG16 and discussed in detail the principles of deep learning and presents the results of experiments and tests. Sharma et al. [36] considered residual network 101. The network parameters adjusted using a method with progressive resizing, with the speed of cyclic learning and the function of focus loss. Zhang et al. [37] made a detailed comparison of VGG and CCD networks for finding weed.

Extremely rare that the same type of weed appears in research that makes it impossible to directly compare several manuscripts. In the works described above, there is not enough argumentation on the choice of models, and the criteria for success are very arbitrary. It can be concluded that, from a practical point of view, these studies are relevant, but do not bring any significant scientific achievements.

## 2.4. Unmanned aerial vehicle for weed recognition tasks

The most promising direction in the use of neural networks in the weed recognation tasks is the possibility of using a small computer system with machine vision function as part of an unmanned aerial vehicle. A few studies describe the implementation process of using an unmanned aerial vehicle for weed control. Since in this case, the quality of the photo/video is limited, due to the unstable position of the aircraft - preprocessing of images is especially important. Huang et al. [38] used the full convolution network (FCN) method to recognize weeds from images collected by the apparatus. Hameed et al. [39], used the aircraft to collect information about the crop, where to classify the structures, a technique was used to train the matrix of coincidences of the gray NN level with Haralix descriptors. Many types of research described a method with a general analysis of productivity without accurate recognition of the crop [40, 41].

Today there are no unmanned aerial vehicles that could overcome the difficulties with recognizing crops on the scale of an agricultural farm and, so far, are usually used to present a general picture of the crop on the field.

A promising area for research in this industry is the use of a single-board computer with aerial vehicles. But today into account that using the deep learning methods on the Raspberry PI due to the limited RAM (1-4 GB) and due to the low processor speed of 1.5 GG is almost impossible ( ResNet> 100 MB, VGGNet> 550 MB, AlexNet> 200 MB, GoogLeNet> 30 MB). Real-time detection with R-CNN, Fast R-CNN, Faster R-CNN, Yolo, RetinaNet has the same recognition speed problem. The solution could be to use - NVIDIA Jetson TX1 and TX2 - a special platform for computing neural networks. The main disadvantage, which is the high cost. Company STMicroelectronics has made it possible to use deep learning methods on microcontrollers. They released X-CUBE-AI - AI has an expansion pack for STM32CubeMX. This extension can work with various deep learning environments such as Caffe, Keras, TensorFlow, Caffe, ConvNetJs, etc. Thanks to this, the neural network can be trained on a desktop computer with the ability to calculate on a GPU. After integration, use the optimized library for the 32-bit STM32 microcontroller. At the same time, using a microcontroller instead of a single-board computer will significantly reduce power consumption by several watts. Therefore, the use of neural networks as part of an unmanned aerial vehicle has good prospects

## 2.5. Development ccompanies in weed recognition tasks

Several companies provide services for the sale of autonomous robots for spot weeding agricultural. lan. Naio-technologies introduced a robot - OZ WEEDING ROBOT. Bilberry Company are developing tracking systems that can be installed on agricultural machinery for subsequent weed monitoring. PrecisionHawk's is developing autonomous aircraft with weed identification systems. Blue River Technology Company displays information on the website on developments in the use of neural networks in agricultural Ecorobotix introduced WEEDING ROBOT. Carre developed the ANATIS robot.

All the companies described above, for the developed technologies, did not provide documentation, the experimental data on which could be used to judge the results of the operation of the devices. At best, companies limited themselves to fragments of a video demonstration of devices. Agro-market is becoming a very popular destination for startups. This is confirmed by the information about large financial investments in various companies engaged in IT development for agricultural enterprises. Working prototypes could not be found. The research results of startups, due to objective reasons, are even more closed than the companies described in the earlier chapter, which is why they are not considered in this study.

## 2.6. Other research in weed recognition tasks

Very popular manuscripts that presented a prototype robotic complex for automatically remove weeds in the fields. For example, Sabanci et al. [21], Young et al. [22], Frasconi et al. [23], Kulkarni et al. [24], Bakhshipour et al. [25], Kargar et al. [26], Potena et al. [27] and Slaughter et al. [28] presented various neural networks for diagnosing weeds and mechanisms to remove weeds. The disadvantage of these works is too much information coverage - mechanics, electronics, neural networks, and robotics. As a result, these works briefly describe implementation of neural networks. Unlike these articles with robotic systems thesis describe as much as possible every aspect of the research. Ospina [29] In the thesis described the development process for smart agricultural vehicle. Ospina used machine vision data as a bifocal imaging device that operates in 2 modes - simultaneous shooting of a wide-angle image and approximate shooting with a telephoto lens. To analyze the presence of weed from a telephoto lens, high-resolution photographs were obtained, and a wide-angle camera allowed increasing the monitoring area. Result with high accuracy obtained in terms of searching for weeds in the field from the camera mounted on the vehicle. But, due to its high price of the camera system of this format is more suitable for research than for subsequent work on the field.

Dyrmann [30], showed in detail image transformation between layer neural models, figure 3. The work uses standard methods of image preprocessing and a convolutional neural network.

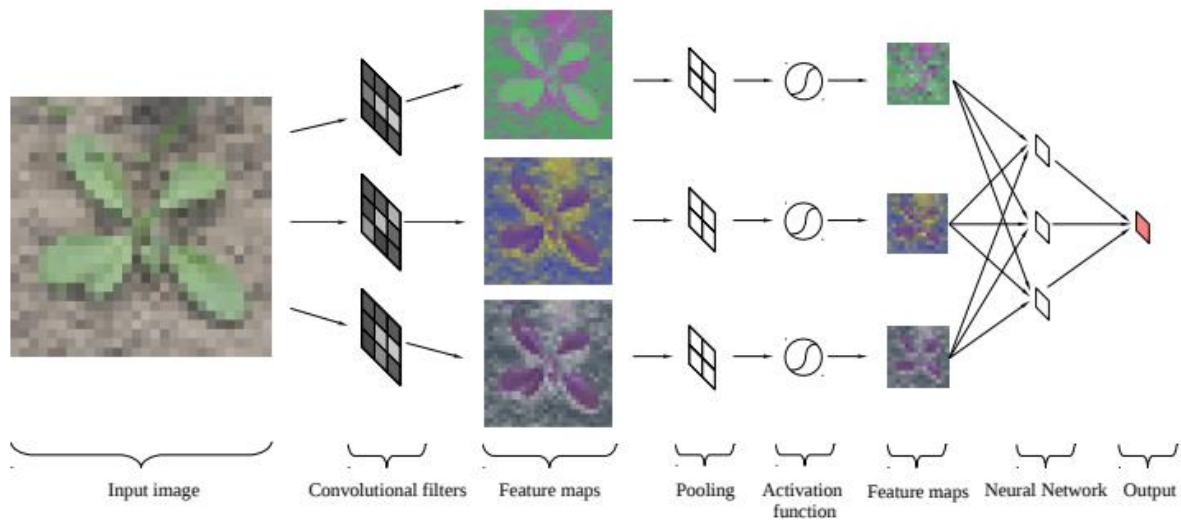

Fig. 3. CNN Model Demonstration

In contrast to this thesis, Alba [31] used not the most popular algorithm based on the FuzzyC-means method for classification problems. In the thesis researched Real-time computer vision technique for robust plant seedling tracking in the field environment. Various segmentation methods were considered.

## 3. Discussion and conclusions

The review showed a wide variety type of neural networks used in the process of weed recognition,

where leading positions were taken by the different implementation of convolutional neural network networks. Based on an analysis of the articles described earlier and the results of Kaggle and Imagenet contests, the following is a recommended list of deep neural networks for classifying images. These models are advisable to use in the weed classification tasks, figure 4.

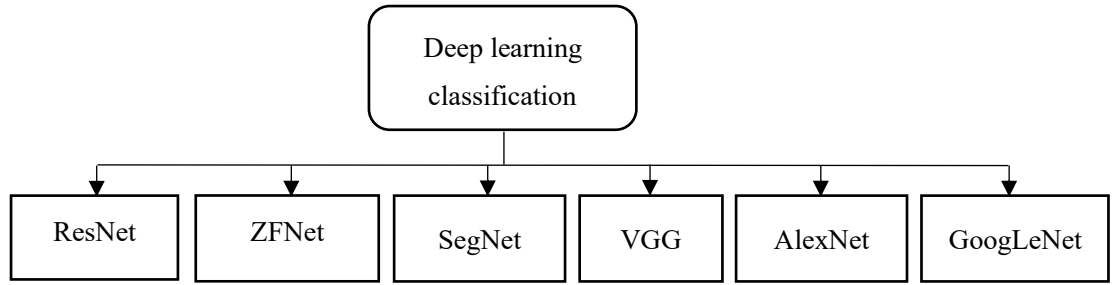

Fig. 4. Neural networks recommended for image classification in plant recognition

For weed recognition in video-stream recommended the next models, figure 5.

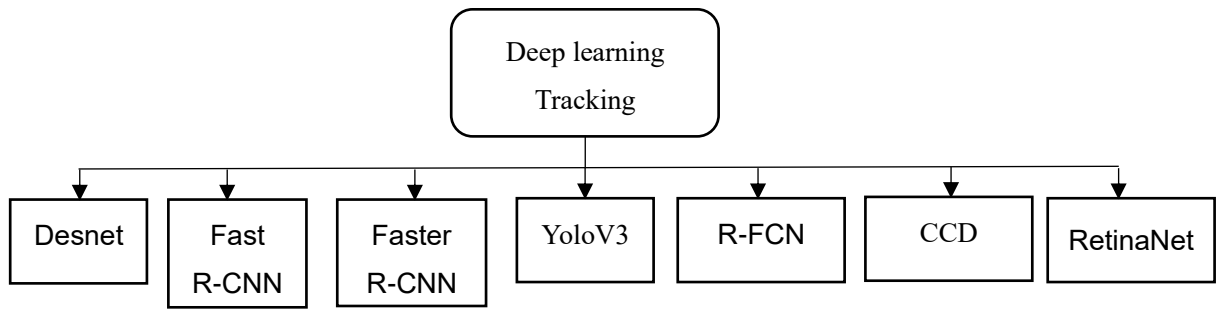

Fig.5. The neural networks that are appropriate to use in the tracking process of weed

The difficulty in building neural networks is that when choosing a model, many parameters depend on the task, the amount of input data, the availability of time and computing power, which makes each work unique. In view of this, it is impossible to correctly model the network initially, but it is necessary to use as many mechanisms as possible in pre-processing and consider a larger number of networks.

A common drawback of all research work is the lack of a full-fledged long-term verification of the neural network in the field with weeds. In the works for verification, the conditions are artificially modeled to obtain a clear image with the plant and weed, which initially simplifies the work for the neural network. The date used for training and testing in open access is not provided, which makes it impossible to analyze the model.

In recognition of weed on the field, two main points can be distinguished. The first is the recognition complexity, which is dictated by weather conditions, the presence of all kinds of options for the various physical locations of weed. The second is the recognition speed and the number of possible objects for detection. In the research described earlier, no links were made to the Kaggle website, which is the best place to analyze the performance of the developed models of the wide network.

Competitions will help to give a real picture of the assessment of machine learning models that are purposefully used in the process of object recognition. Also, research does not come up with links to image.net. In this case, the speed of recognition of objects is not disclosed in the works. In order to increase the recognition speed and increase the number of tracking objects, in addition to increasing the computer power, it is necessary to apply non-typical actions like using two GPUs in the stereo vision algorithm or using several GPUs in the detection algorithm.

**Conflicts of Interest**: None

**Funding**: None

**Ethical Approval**: Not required